\documentclass[runningheads]{llncs}

\usepackage{eccvabbrv}

\usepackage{graphicx}
\usepackage{booktabs}

\usepackage[accsupp]{axessibility}  % Improves PDF readability for those with disabilities.

\usepackage{hyperref}
\usepackage{url}
\usepackage{kotex}
\usepackage{wrapfig}
\usepackage{multirow}
\usepackage{colortbl}
\usepackage{array}
\usepackage[table]{xcolor}
\usepackage{amsmath}
\usepackage{siunitx}
\usepackage{enumitem}
\usepackage{subcaption}
\usepackage[export]{adjustbox}
\usepackage{placeins}
\usepackage{float}
\definecolor{gtgreen}{HTML}{2ECC71}    
\definecolor{hallured}{HTML}{E74C3C}     
\definecolor{markyellow}{HTML}{FFD166}  
\definecolor{marklilac}{HTML}{E3D0FF} 
\usepackage{orcidlink}

\begin{document}

\title{Structured Redundancy Modeling for Efficient Visual Token Pruning in High-Resolution MLLMs} 

\titlerunning{Structured Redundancy Modeling for Efficient Visual Token Pruning}
\authorrunning{J. Song et al.}

\author{
Juwon Song\inst{1}
\and
Woohyeong Kim\inst{2}
\and
Kyeongbo Kong\inst{2}\thanks{Corresponding author.}
}

\authorrunning{J. Song et al.}

\institute{
LG Electronics, Seoul, Republic of Korea\\
\email{wn5649@sogang.ac.kr}
\and
Pusan National University, Busan, Republic of Korea\\
\email{\{zovw8179, kbkong\}@pusan.ac.kr}\\
\vspace{2pt}
\url{https://github.com/cvsp-lab/SFPruner}
}

% % TODO FINAL: Replace with your institution list.
% Springer Heidelberg, Tiergartenstr.~17, 69121 Heidelberg, Germany

\maketitle

\begin{abstract}
Recent high-resolution Multimodal Large Language Models (MLLMs) generate thousands of visual tokens per input, leading to a visual token explosion that introduces severe latency bottlenecks. While token pruning mitigates this issue, state-of-the-art subset-optimization methods typically rely on iterative subset construction to jointly capture visual diversity and instruction relevance. As visual token counts scale, this sequential dependency introduces significant selection overhead, severely limiting the translation of theoretical FLOPs reductions into actual wall-clock speedups. To address this limitation, we propose Single-Forward Pruner (SFPruner), a structural reformulation of visual token pruning that embeds redundancy control directly into the scoring space, bypassing the need for iterative combinatorial optimization. Our non-iterative framework achieves redundancy-aware importance selection in a single forward pass through two complementary mechanisms. First, to attenuate redundancy at the covariance level, we introduce a semantics-guided ridge leverage scheme. By integrating instruction relevance and visual saliency, this mechanism suppresses dominant covariance directions and mitigates representation bias. Second, ranking-based directional masking resolves residual overlap through asymmetric similarity competition, where higher-scoring tokens explicitly suppress redundant lower-scoring alternatives via parallel tensor operations. Extensive evaluations demonstrate that our approach maintains stable selection costs, reducing the token selection process by up to 110 ms (from 112.4 ms to just 2.5 ms at 512 tokens in Qwen2.5-VL). This structural efficiency successfully translates theoretical token reductions into tangible inference speedups while preserving highly competitive performance against state-of-the-art techniques under aggressive compression.

\end{abstract}

\section{Introduction}

Recent multimodal large language models (MLLMs)~\cite{liu2023visual,wang2024qwen2,liu2024deepseek} extend beyond text processing to perform fine-grained visual reasoning over high-resolution images and long video sequences. Architectures like LLaVA-NeXT \cite{liu2024llavanext} and Qwen-VL \cite{wang2024qwen2} generate thousands of visual tokens per input using dynamic-resolution mechanisms. However, since self-attention scales quadratically with token length, this leads to severe latency and memory bottlenecks, limiting practical deployment.

\begin{figure*}[!t]
    \centering
    \includegraphics[width=0.9\linewidth]{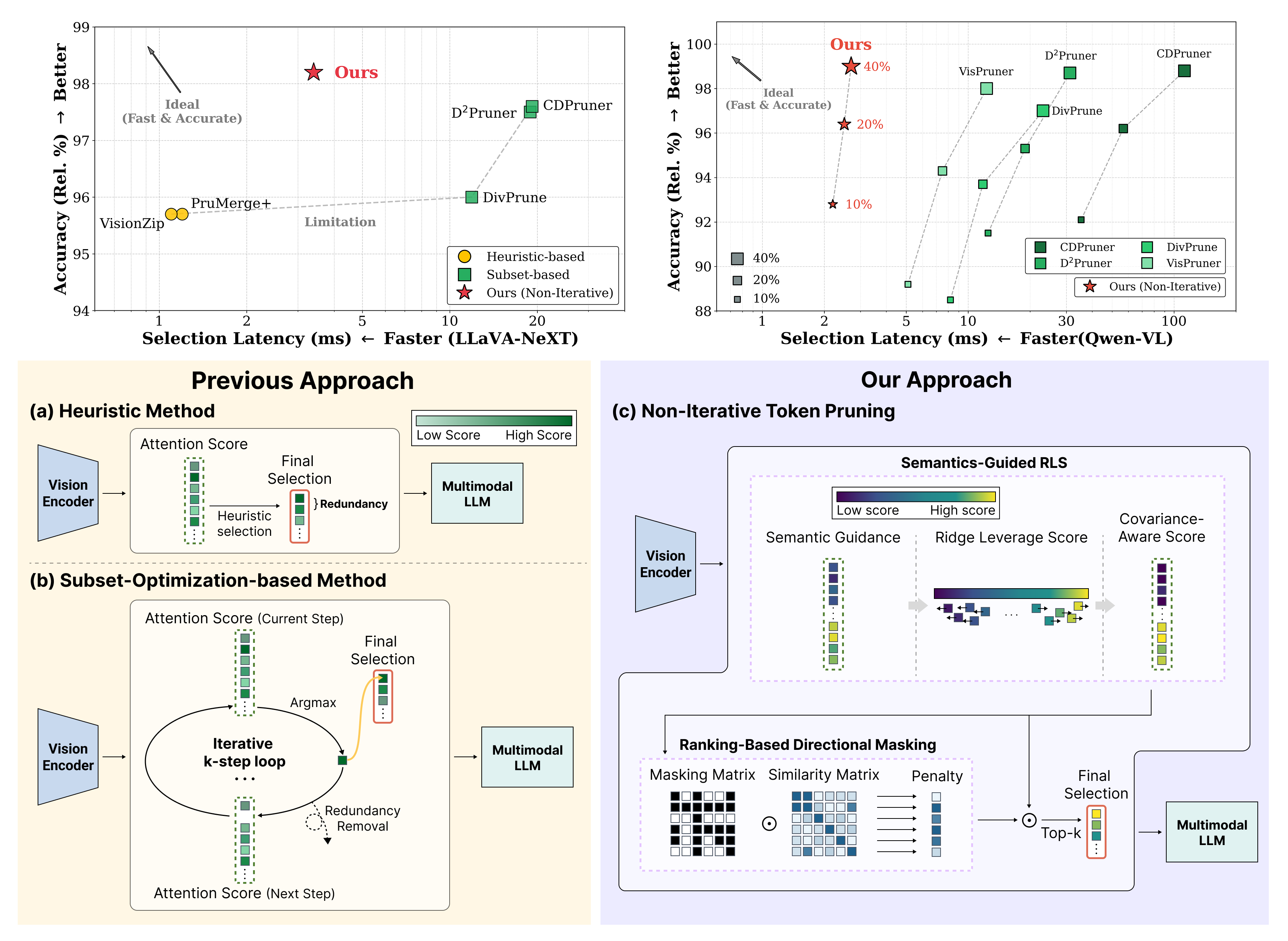}
    \vspace{-3mm}
    \caption{\textbf{Conceptual comparison of token selection frameworks.} \textbf{(a)} Heuristic (Top-$K$) relies on isolated scores, lacking structural redundancy control. \textbf{(b)} Traditional sequential search mitigates redundancy but introduces iterative dependencies. \textbf{(c)} Our framework enables efficient single-pass selection via SG-RLS and directional masking, eliminating iteration overhead.}
    \vspace{-5mm}
    \label{fig:teaser}
\end{figure*}

Vision token pruning addresses the challenge of computational efficiency by selecting informative tokens and removing redundant ones prior to costly attention operations~\cite{xing2024pyramiddrop,fastv,zhang2024sparsevlm}. Existing approaches can be broadly categorized into two groups, each exhibiting a distinct trade-off between speed and performance, as illustrated in the top panels of Fig.~\ref{fig:teaser}. Recent empirical evidence further supports this dichotomy: attention-based pruning tends to be effective for simple images with concentrated visual evidence, whereas diversity-oriented pruning is more beneficial for complex images with distributed visual cues~\cite{baek2026agilepruner}. Heuristic methods (Fig.~\ref{fig:teaser}a) employ ranking or filtering strategies based on attention or similarity signals~\cite{vispruner,shang2025prumerge,yang2025visionzip}. As shown in the top-left panel, these methods offer high speed but fail to fully eliminate spatial redundancy and often experience substantial performance degradation under aggressive pruning. In contrast, subset-optimization methods (Fig.~\ref{fig:teaser}b) explicitly model both importance and diversity to better preserve accuracy, using formulations such as DPP, diversity maximization, or graph-based MWIS~\cite{zhang2025cdpruner,alvar2025divprune,Song_2026_CVPR}. However, these approaches depend on iterative selection procedures, resulting in computational overhead that increases significantly with the number of tokens. This limitation is evident in the top-right panel of Fig.~\ref{fig:teaser}, where the latency of iterative subset-optimization methods rises sharply as the token retention ratio increases.

In this work, we propose Single-Forward Pruner (SFPruner), a structural approach to token pruning (\textbf{Fig.~\ref{fig:teaser}c}). Instead of iterative subset construction, we embed redundancy control directly into the scoring process. Our SFPruner decomposes redundancy handling into two mechanisms in a single forward pass. First, we apply \textbf{covariance-level redundancy attenuation} via a semantics-guided ridge leverage scheme~\cite{r1,r2}, integrating instruction relevance and visual saliency to modulate the structural scores. This step suppresses dominant features and regularizes scores. Second, we use \textbf{ranking-based directional masking}, where higher scores suppress redundant lower ones through asymmetric similarity competition. This resolves token overlap without iterative updates, enabling direct top-$k$ selection.

By controlling redundancy at both covariance and pairwise levels, our approach enables efficient, redundancy-aware selection in a single forward pass. Leverage-based reweighting mitigates dominance bias, and directional masking enforces structured competition. As token counts grow, our non-iterative design avoids the sequential bottlenecks of previous methods. As demonstrated in Fig.~\ref{fig:teaser}, our method maintains constant, minimal selection costs, reducing latency from 112.4~ms to just 2.5~ms when selecting 512 tokens in Qwen2.5-VL \cite{bai2025qwen}. This ensures that FLOP reductions translate directly into real-world speedup without sacrificing performance, even under aggressive pruning.

\noindent\textbf{Our main contributions are:} 
\begin{itemize} 
\item \textbf{A structural reformulation of visual token pruning:} We show that redundancy-aware selection can be achieved in feature space without combinatorial optimization, by embedding redundancy control into the scoring process. 
\item \textbf{Covariance-level redundancy attenuation:} We introduce a semantics-guided leverage reweighting mechanism that integrates instruction relevance and visual saliency, suppressing dominant covariance directions before selection.
\item \textbf{Ranking-based directional masking:} We propose an asymmetric masking strategy, where higher-scoring tokens suppress redundant lower-scoring tokens in a single tensor operation, removing the need for iterative subset construction. 
\item \textbf{Scalable efficiency for high-resolution inputs:} Our method maintains stable selection costs. Reducing selection latency from 112.4~ms to 2.5~ms translates theoretical FLOP reductions into practical acceleration, preserving highly competitive reasoning capability (retaining up to 99.0\% relative performance).
\end{itemize}

\section{Related Works}
\label{sec:related_works}

\subsection{Vision Token Pruning Methods}

Pruning techniques that reduce visual token redundancy are widely adopted to enhance the inference efficiency of Multimodal Large Language Models (MLLMs), largely because they require no additional training. Recent approaches broadly fall into two categories: heuristic methods, which rely on direct scoring signals, and subset-optimization-based methods, which explicitly model relationships between tokens.
% \textbf{\cite{DART, shang2025prumerge, FasterVLM, fastv, zhang2024sparsevlm}}

\noindent\textbf{Heuristic Selection and Merging:}
Numerous attention-based methods select tokens relying strictly on local attention weights, often ignoring semantic redundancy \cite{DART,shang2025prumerge,FasterVLM,fastv,zhang2024sparsevlm}. Attempts to extract global features \cite{arif2025hired} or dynamically adjust token counts \cite{ye2025atp} tend to favor dominant objects, risking the loss of fine-grained cues. Merging techniques such as ToMe~\cite{bolya2023tome} and VisionZip \cite{yang2025visionzip} reduce redundancy through token merging, aggregation, or recognition-oriented importance scoring. However, these vision-only compression strategies are primarily designed for generic visual redundancy reduction and may obscure fine-grained visual details critical for instruction-conditioned reasoning in high-resolution MLLMs. In contrast, MLLM token pruning requires preserving query-relevant visual evidence while suppressing redundant or task-irrelevant tokens.

\noindent\textbf{Subset-Optimization-Based Token Selection:} To overcome these limitations, recent works formulate token selection as combinatorial optimization—e.g., maximizing importance and diversity via DPP~\cite{zhang2025cdpruner}, MMDP~\cite{alvar2025divprune}, or graph-based MWIS formulations~\cite{Song_2026_CVPR}. As these problems are NP-hard, \textbf{sequential greedy search} is universally adopted for polynomial-time approximation. This requires an iterative $k$-step loop to update states after each selection. While effective for redundancy-aware selection, this sequential paradigm introduces a severe latency bottleneck that scales linearly with sequence length in high-resolution settings.

\subsection{Ridge Leverage Score in Neural Architectures}

Based on Randomized Numerical Linear Algebra (RandNLA) \cite{RLA}, the Ridge Leverage Score (RLS) is a statistical metric that balances dominant patterns and sparse details by softening linear dependencies within data through ridge regularization ($\lambda$) \cite{r1,r2}. Notably, RLS provides the mathematical advantage of simultaneously quantifying the global diversity and feature-space uniqueness of all data points via matrix inversion, thereby circumventing the need for iterative state updates. Due to these structural properties, RLS has become widely adopted for dimensionality reduction and computational optimization in deep learning architectures. In Natural Language Processing (NLP), it has been used to compress the effective dimensions of transformer blocks, as demonstrated in MoDeGPT \cite{r3}. Furthermore, a recent attention approximation study \cite{song2026sublinear} introduced RLS sampling to quantum algorithmic environments, reducing the standard $\mathcal{O}(N^2)$ attention complexity to sublinear time. As such, RLS has emerged as a core technique for extracting principal components without compromising the information spectrum.

\begin{figure*}[t]
    \centering
    \includegraphics[width=0.95\linewidth]{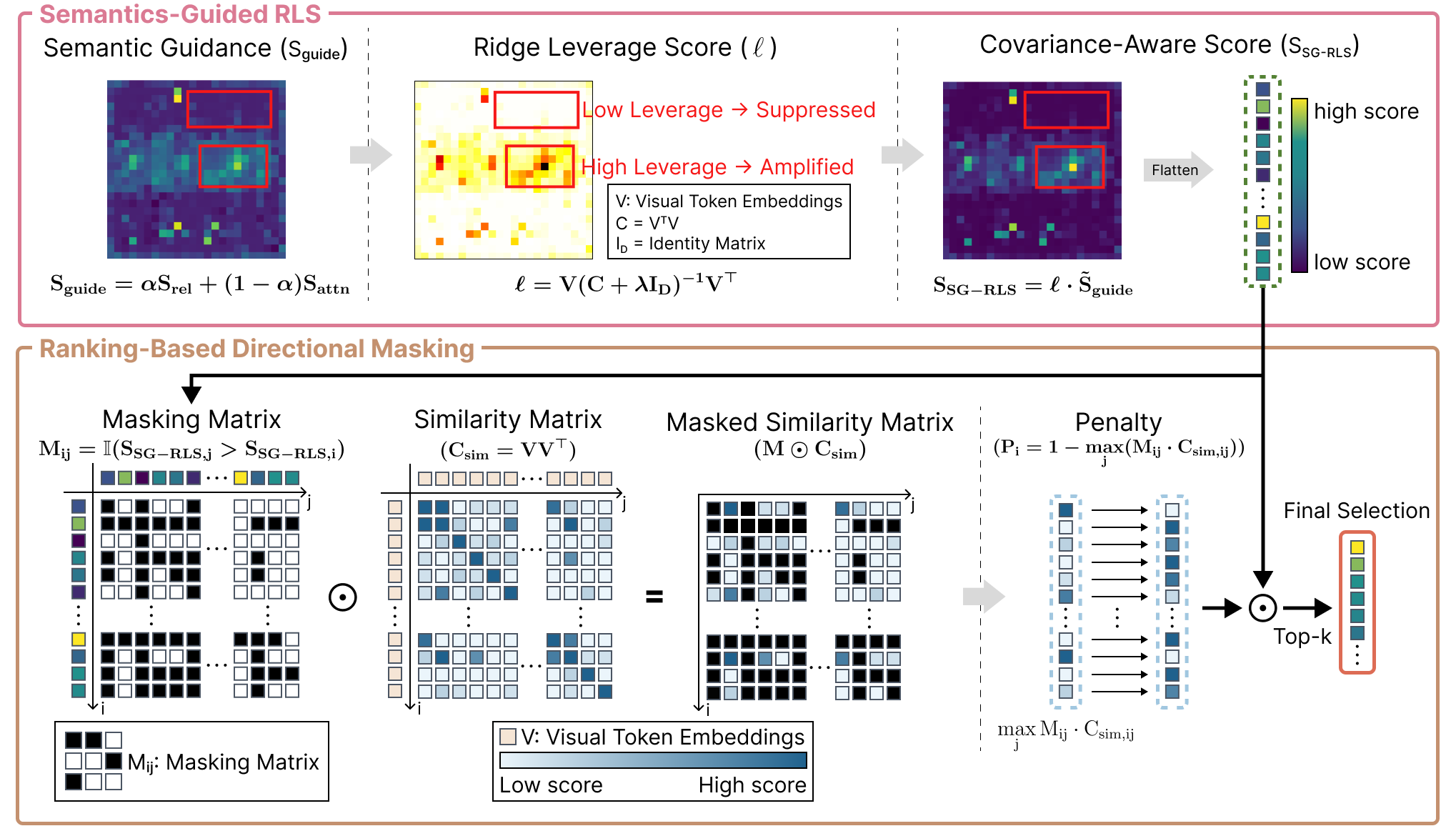}
    \vspace{-2mm}
    \caption{\textbf{Pipeline of the non-iterative token pruning framework.}  \textbf{(Top) Semantics-Guided RLS (SG-RLS)}: The semantic guidance ($S_{\mathrm{guide}}$) is modulated by the leverage score map ($\ell$) to suppress redundant background features, yielding a covariance-aware score ($S_{\mathrm{SG\text{-}RLS}}$). \textbf{(Bottom) Ranking-Based Directional Masking}: A ranking mask ($M$) is applied to the pairwise similarity matrix. Tokens receive a directional suppression penalty ($P_i$) based on their maximum similarity to higher-priority competitors, extracting the final pruned subset through single-pass tensor operations.}
    \label{fig:main_arch}
    \vspace{-5mm}
\end{figure*}

\section{Proposed Method: Single-Forward Pruner (SFPruner)}
\label{sec:method}

In this section, we present SFPruner, a non-iterative token pruning framework that bypasses sequential subset construction by embedding redundancy modeling directly into the scoring process. Our method consists of three key steps:
(1) establishing base token importance via instruction-aware semantic guidance (\S\ref{subsec:prior});
(2) attenuating global covariance-level redundancy using a semantics-guided ridge leverage formulation (\S\ref{subsec:rls}); and
(3) resolving residual local overlap via ranking-based directional masking, where higher-scoring tokens asymmetrically suppress correlated lower-scoring alternatives (\S\ref{subsec:masking}).

\subsection{Semantic Guidance: Textual Relevance and Visual Saliency}
\label{subsec:prior}

We first estimate a base importance score for each visual token, referred to as the semantic guidance. 
This guidance integrates instruction relevance and intrinsic visual saliency, 
providing an instruction-aware base score that is subsequently modulated by covariance-level redundancy modeling.

Let $V \in \mathbb{R}^{N \times D}$ denote the matrix of $L_2$-normalized visual token embeddings, 
where each row $v_i \in \mathbb{R}^{1 \times D}$ corresponds to a visual token. 
Let $q \in \mathbb{R}^{1 \times D}$ denote the projected text embedding.

\paragraph{\textbf{Textual relevance.}}
We measure instruction relevance using cosine similarity between each visual token and the text embedding, followed by temperature-scaled normalization:

\begin{equation}
S_{\mathrm{rel},i}
=
\frac{
\exp\!\left(
\tau \, \frac{v_i q^\top}{\|v_i\|\,\|q\|}
\right)
}{
\sum_{j=1}^{N}
\exp\!\left(
\tau \, \frac{v_j q^\top}{\|v_j\|\,\|q\|}
\right)
},
\end{equation}
where $\tau$ controls the sharpness of the relevance distribution.

\paragraph{\textbf{Visual saliency.}}
To capture intrinsic visual importance independent of the instruction, 
we extract a saliency score from the vision encoder self-attention. 
Specifically, we use the attention weight from the global token (e.g., CLS) to each visual token:

\begin{equation}
S_{\mathrm{attn},i}
=
\mathrm{Attention}(\mathrm{CLS} \rightarrow v_i).
\end{equation}

In the absence of a canonical CLS token (e.g., Qwen2.5-VL \cite{bai2025qwen}), we construct a proxy for global saliency by aggregating the self-attention weights across all spatial tokens.

\paragraph{\textbf{Semantic guidance fusion.}}
We combine textual relevance and visual saliency through a linear fusion:

\begin{equation}
S_{\mathrm{guide},i}
=
\alpha S_{\mathrm{rel},i}
+
(1-\alpha) S_{\mathrm{attn},i},
\quad \alpha \in [0,1].
\end{equation}

For architectures lacking a dedicated text encoder (e.g., Qwen2.5-VL \cite{bai2025qwen}), we rely solely on the visual saliency as the semantic guidance. To ensure numerical stability and comparable scaling across inputs, 
we apply Min-Max normalization:

\begin{equation}
\tilde{S}_{\mathrm{guide},i}
=
\frac{
S_{\mathrm{guide},i}
-
\min_j S_{\mathrm{guide},j}
}{
\max_j S_{\mathrm{guide},j}
-
\min_j S_{\mathrm{guide},j}
+
\epsilon
}.
\label{eq:prior_fusion}
\end{equation}

The normalized semantic guidance $\tilde{S}_{\mathrm{guide}}$ serves as the base importance score for the covariance-level redundancy attenuation described in Sec.~\ref{subsec:rls}.

\subsection{Covariance-Level Attenuation: Semantics-Guided Ridge Leverage Score}
\label{subsec:rls}

Sequential subset selection methods suppress redundancy 
by progressively updating the residual feature space 
after each token is selected. 
While effective, such stepwise dependency introduces 
non-trivial execution overhead.
We instead perform redundancy attenuation globally 
through a ridge-regularized leverage formulation, 
eliminating the need for iterative updates.

We define the feature covariance matrix

\begin{equation}
C = V^\top V \in \mathbb{R}^{D \times D}.
\end{equation}

The ridge leverage score of token $i$ is defined as

\begin{equation}
\ell_i
=
v_i
\left(
C + \lambda I_D
\right)^{-1}
v_i^\top,
\label{eq:rls}
\end{equation}
where $\lambda$ is a small regularization constant.

The inverse covariance $(C + \lambda I_D)^{-1}$ 
attenuates dominant eigendirections of the feature space, 
reducing the influence of globally redundant patterns. 
Tokens aligned with high-energy subspaces 
receive lower leverage scores, 
while structurally distinctive tokens obtain higher scores.

To incorporate task-awareness, 
we modulate the structural score with the semantic guidance introduced in Section~\ref{subsec:prior}. 
The final covariance-level redundancy score is defined as
\begin{equation}
S_{\mathrm{SG\text{-}RLS},i}
=
\ell_i
\cdot
\tilde{S}_{\mathrm{guide},i}.
\end{equation}

The multiplicative formulation acts as a soft-AND gate, requiring both structural uniqueness and semantic relevance for a token to receive a high score. Consequently, tokens that are structurally distinctive but task-irrelevant, or semantically relevant but highly redundant, are naturally down-weighted. This formulation separates geometric uniqueness from semantic importance: ridge leverage captures global structural redundancy, while the semantic guidance suppresses task-irrelevant outliers.

Importantly, the entire computation consists of parallel matrix operations and a single matrix inversion. While the covariance matrix $C$ is intrinsically $D \times D$, we leverage linear algebraic duality via the Woodbury matrix identity to establish an equivalence between the feature covariance and the $N \times N$ Gram matrix. This duality allows us to flexibly perform the inversion on the smaller dimension between the sequence length $N$ and the feature dimension $D$. Consequently, we can circumvent the computationally prohibitive $N \times N$ inversion in high-resolution regimes ($N > D$) while completely removing iterative dependencies. Detailed derivations for this identity are provided in the supplementary material.

\subsection{Ranking-Based Directional Masking}
\label{subsec:masking}

While SG-RLS mitigates redundancy at the covariance level, subset-based optimization methods further impose pairwise exclusion: once a token is selected, highly similar alternatives are suppressed.
We incorporate this pairwise redundancy control 
without introducing iterative updates.

Let $C_{\mathrm{sim}} = VV^\top \in \mathbb{R}^{N \times N}$
denote the cosine similarity matrix between tokens. Higher values indicate stronger semantic overlap.

We define a directional masking mechanism 
based on the covariance-level scores 
$S_{\mathrm{SG\text{-}RLS}}$ from Section~\ref{subsec:rls}.
For each token $i$, suppression is applied only 
from tokens with strictly higher scores.
Formally, we construct a masking matrix

\begin{equation}
M_{ij}
=
\mathbb{I}\!\left(
S_{\mathrm{SG\text{-}RLS},j}
>
S_{\mathrm{SG\text{-}RLS},i}
\right),
\end{equation}
where $\mathbb{I}(\cdot)$ denotes the indicator function.
This mask enforces asymmetric competition: 
higher-ranked tokens may suppress lower-ranked ones,
but not vice versa.

The directional suppression penalty for token $i$ is then

\begin{equation}
P_i
=
1 -
\max_j
\left(
M_{ij} \cdot C_{\mathrm{sim},ij}
\right).
\label{eq:penalty}
\end{equation}
% By construction, (M{ii} = 0), ensuring the maximum is always non-negative. 
% This guarantees that negatively correlated tokens cannot inadvertently increase the score, i.e., ($P_i \le 1$).
By construction, $M_{ii}=0$, ensuring that the maximum is non-negative. 
This prevents negatively correlated tokens from increasing the score, i.e., $P_i \le 1$.
If no higher-scoring token exhibits strong similarity,
$P_i$ remains close to 1; 
otherwise, the penalty increases proportionally 
to the maximum overlap. The final redundancy-aware score is computed as
\begin{equation}
S_{\mathrm{final},i}
=
S_{\mathrm{SG\text{-}RLS},i}
\cdot
P_i.
\end{equation}

This formulation enforces structured pairwise suppression 
through fully parallel tensor operations.
All computations consist of matrix multiplications, 
element-wise comparisons, and row-wise reductions,
eliminating iterative subset updates 
while maintaining directional redundancy control.
The final subset is obtained via a single Top-$K$ operation.

\section{Experiments}
\subsection{Experimental Setup}

We evaluate the effectiveness and structural efficiency of the proposed non-iterative token pruning framework across multiple Multimodal Large Language Model (MLLM) architectures and diverse vision-language benchmarks.
\begin{itemize}
\item \textbf{Models.}
To demonstrate generality, we evaluate three representative MLLM families:
LLaVA-NeXT-7B~\cite{liu2024llavanext}, optimized for high-resolution image processing;
Qwen2.5-VL-7B~\cite{bai2025qwen}, a recent high-capacity vision-language model; and
LLaVA-Video-7B~\cite{zhang2024videoinstructiontuningsynthetic}, which supports multi-frame video inference.
These models cover both image-centric and temporal multimodal settings.

\item \textbf{Benchmarks.}
For image understanding, we evaluate on VQAv2~\cite{goyal2017vqav2}, GQA~\cite{hudson2019gqa}, TextVQA~\cite{textvqa}, MME~\cite{mme}, and ScienceQA (SQA)~\cite{ScienceQA}.
For video reasoning, we use VideoMME~\cite{fu2025video}, MLVU~\cite{zhou2024mlvu}, and LongVideoBench~\cite{wu2024longvideobench}.
These benchmarks collectively test fine-grained reasoning, instruction alignment, and large-scale visual token processing.

\item \textbf{Evaluation Metrics.}
We report absolute task accuracy and relative performance retention (Rel.~\%) with respect to the corresponding unpruned vanilla models.
To evaluate computational behavior, we measure the wall-clock Selection Latency (ms), which isolates the execution time of the token selection module prior to the LLM forward pass. The reported latency includes the complete pruning pipeline (semantic guidance, SG-RLS scoring, directional masking, and Top-$K$ selection), while excluding the vision encoder and LLM forward pass. All latency measurements are performed on a single NVIDIA RTX 4090 GPU with synchronized CUDA timing to ensure consistent profiling.
\end{itemize}
Detailed parameter configurations ($\alpha$ and $\lambda$) and additional sensitivity analyses are provided in the supplementary material.

\begin{table*}[t]
\centering
\small
\vspace{1mm}
\resizebox{0.95\textwidth}{!}{
\renewcommand{\arraystretch}{1.1}
\begin{tabular}{lccccccccccc} 
\toprule
\textbf{Method} &
\textbf{VQA\textsuperscript{v2}} &
\textbf{GQA} &
\textbf{SQA\textsuperscript{IMG}} &
\textbf{TextVQA} &
\textbf{POPE} &
\textbf{MME} &
\textbf{MMB} &
\textbf{MMB\textsuperscript{CN}} &
\textbf{MMVet} &
\textbf{Rel. (\%)} &
\textbf{Time (ms)} $\downarrow$ \\
\midrule
\rowcolor{orange!8}
\multicolumn{12}{c}{\emph{Vanilla $5 \times 576$ Tokens}} \\
LLaVA-NeXT-7B & 81.3 & 62.5 & 67.5 & 60.3 & 86.8 & 1511.8 & 65.8 & 57.3 & 40.0 & 100.0 & 0.00 \\
\midrule
\rowcolor{orange!8}
\multicolumn{12}{c}{\emph{Retain $5 \times 128$ Tokens}} \\
\rowcolor{blue!5} SparseVLM (ICML'25) \cite{zhang2024sparsevlm} & 79.2 & 61.2 & 67.6 & 59.7 & 85.3 & 1456.8 & 65.9 & \textbf{58.6} & 36.1 & 97.9 & 19.0 \\
\rowcolor{blue!5} PruMerge+ (ICCV'25) \cite{shang2025prumerge}  & 78.2 & 60.8 & 67.8 & 54.9 & 85.3 & 1480.2 & 64.6 & 57.3 &32.7 & 98.3 & 1.2 \\
\rowcolor{blue!5} VisionZip (CVPR'25) \cite{yang2025visionzip} & 79.1 & 61.2 & \textbf{68.1} & \textbf{59.9} & 86.0 & 1493.4 & 65.8 & 58.1 & 38.9 & 99.5 & 1.1 \\
\rowcolor{green!5} DivPrune (CVPR'25) \cite{alvar2025divprune} & 79.3 & 61.9 & 67.8 & 57.0 & 86.9 & 1469.7 & 65.8 & 57.3 & 38.0 & 98.5 & 33.8 \\
\rowcolor{green!5} CDPruner (NeurIPS'25) \cite{zhang2025cdpruner} & 79.9 & 62.6 & 67.9 & 58.4 & \textbf{87.3} & 1474.2 & 66.3 & 57.5 & \textbf{41.9} & 99.9 & 35.2 \\
\rowcolor{green!5} \textbf{SFPruner} (Ours) & \textbf{80.1} & \textbf{62.7} & 67.6 & 59.0 & 87.1 & \textbf{1480.3} & \textbf{67.1} & 57.6 & 41.0 & \textbf{100.0} & \textbf{3.5} \\
\midrule
\rowcolor{orange!8}
\multicolumn{12}{c}{\emph{Retain $5 \times 64$ Tokens}} \\
\rowcolor{blue!5} SparseVLM (ICML'25) \cite{zhang2024sparsevlm} & 74.6 & 57.9 & 67.2 & 56.5 & 76.9 & 1386.1 & 63.1 & \textbf{56.7} & 32.8 & 93.3 & 18.2 \\
\rowcolor{blue!5} PruMerge+ (ICCV'25) \cite{shang2025prumerge} & 75.3 & 58.8 & \textbf{68.1} & 54.0 & 79.5 & 1444.2 & 63.0 & 55.6 & 31.4 & 95.7 & 1.2 \\
\rowcolor{blue!5} VisionZip (CVPR'25) \cite{yang2025visionzip} & 76.2 & 58.9 & 67.5 & \textbf{58.8} & 82.3 & 1397.1 & 63.3 & 55.6 & 35.8 & 95.7 & 1.1 \\
\rowcolor{green!5} DivPrune (CVPR'25) \cite{alvar2025divprune} & 77.2 & 61.1 & 67.7 & 56.2 & 84.7 & 1423.3 & 63.9 & 55.7 & 34.8 & 96.0 & 11.9 \\
\rowcolor{green!5} CDPruner (NeurIPS'25) \cite{zhang2025cdpruner} & 78.4 & \textbf{61.6} & 67.8 & 57.4 & \textbf{87.2} & 1453.1 & \textbf{65.5} & 55.7 & \textbf{37.9} & 97.6 & 19.2 \\
\rowcolor{green!5} \textbf{SFPruner} (Ours) & \textbf{78.5} & 61.5 & 67.8 & 58.3 & 86.7 & \textbf{1507.6} & 65.1 & 56.6 & \textbf{37.9} & \textbf{98.2} & \textbf{3.4} \\
\midrule
\rowcolor{orange!8}
\multicolumn{12}{c}{\emph{Retain $5 \times 32$ Tokens}} \\
\rowcolor{blue!5} PruMerge (ICCV'25) \cite{shang2025prumerge} & 70.5 & 56.2 & 66.9 & 50.3 & 71.1 & 1289.6 & 58.0 & 48.9 & 29.3 & 87.7 & 1.1 \\
\rowcolor{blue!5} VisionZip (CVPR'25) \cite{yang2025visionzip} & 71.4 & 55.2 & \textbf{67.9} & 55.0 & 74.9 & 1327.8 & 58.6 & 50.4 & 32.3 & 90.0 & 1.1 \\
\rowcolor{green!5} DivPrune (CVPR'25) \cite{alvar2025divprune} & 75.0 & 59.3 & 67.1 & 54.1 & 80.0 & 1356.6 & 62.9 & 53.7 & 32.0 & 92.9 & 7.8 \\
\rowcolor{green!5} CDPruner (NeurIPS'25) \cite{zhang2025cdpruner} & \textbf{76.7} & \textbf{60.8} & 67.5 & 55.4 & \textbf{86.8} & 1425.1 & 64.2 & 53.8 & 36.2 & 95.5 & 8.9 \\
\rowcolor{green!5} \textbf{SFPruner} (Ours) & 76.6 & 60.3 & \textbf{67.9} & \textbf{57.0} & 86.6 & \textbf{1435.3} & \textbf{64.3} & \textbf{55.0} & \textbf{37.2} & \textbf{96.4} & \textbf{3.4} \\
\bottomrule
\end{tabular}}
\vspace{1mm}
\caption{Performance comparison on LLaVA-NeXT-7B at different token retention budgets. Relative performance (\textbf{Rel.}) is normalized to the unpruned vanilla model. Light blue rows indicate heuristic-based methods, and light green rows indicate optimization-based methods.}
\label{tab:highres}
\vspace{-5mm}
\end{table*}

%\subsection{Results and Analysis}

\subsection{Performance on LLaVA-NeXT-7B: High-Resolution Multi-Patch Setting}
We evaluate SFPruner on LLaVA-NeXT-7B \cite{liu2024llavanext} under high-resolution settings.
Using the dynamic AnyRes strategy, each image is divided into up to five patches,
yielding 2,880 visual tokens ($5 \times 576$).
To analyze performance under constrained budgets,
we progressively reduce the 576 tokens per patch to 128, 64, and 32 tokens,
corresponding to total retention budgets of 640, 320, and 160 tokens.
Table~\ref{tab:highres} reports task accuracy, relative performance retention (Rel.\%), 
and selection latency.

\paragraph{\textbf{Selection Efficiency.}}
Subset-optimization methods such as DivPrune and CDPruner
rely on sequential search,
resulting in noticeable selection overhead
(e.g., 33.8\,ms and 35.2\,ms at 640-token retention).
In contrast, SFPruner performs covariance-level attenuation
and directional masking via parallel tensor operations,
achieving a stable selection latency of 3.5\,ms across retention budgets.
Importantly, the selection latency remains nearly constant
as the number of retained tokens decreases,
indicating the absence of iterative dependency.

\paragraph{\textbf{Performance Retention.}}
At 640-token retention (approximately 22\% of the original tokens),
our method achieves 100.0\% relative performance,
matching or slightly exceeding the unpruned baseline within evaluation variance.
At 320-token retention (11\% of tokens),
performance remains at 98.2\%,
comparable to optimization-based methods while requiring substantially lower selection time.
Under more aggressive compression (160 tokens, 5.5\% retention),
our method retains 96.4\% performance,
remaining competitive with CDPruner (95.5\%)
while maintaining significantly lower selection latency.
Compared to heuristic approaches,
which exhibit larger performance drops under high compression,
the proposed framework maintains stable reasoning accuracy.
At the same time, it achieves efficiency close to lightweight heuristics,
without relying on sequential subset construction.

\begin{table*}[t]
\centering
\small
\vspace{1mm}
\resizebox{0.95\textwidth}{!}{
\renewcommand{\arraystretch}{1.1}
\begin{tabular}{lcccccccccc}
\toprule
\textbf{Method} &
\textbf{TextVQA} &
\textbf{AI2D} &
\textbf{HBench} &
\textbf{MME} &
\textbf{MMB} &
\textbf{MMB\textsuperscript{CN}} &
\textbf{POPE} &
\textbf{MMStar} &
\textbf{Rel. (\%)} &
\textbf{Time (ms)} $\downarrow$ \\
\midrule
\rowcolor{orange!8}
\multicolumn{11}{c}{\emph{Vanilla 100\% Tokens}} \\
Qwen2.5-VL-7B & 85.3 & 80.8 & 64.3 & 2316.0 & 79.7 & 81.8 & 86.4 & 56.7 & 100.0 & - \\

\midrule
\rowcolor{orange!8}
\multicolumn{11}{c}{\emph{Retain 40\% Tokens (Avg. 512 Tokens)}} \\
\rowcolor{blue!5} PruMerge (ICCV'25) \cite{shang2025prumerge} & 82.1 & 79.8 & 62.6 & 2287.0 & 78.4 & 80.0 & 86.0 & 55.6 & 98.1 & \textbf{1.2} \\
\rowcolor{blue!5} VisPruner (ICCV'25) \cite{vispruner} & 82.6 & 79.9 & 61.7 & 2297.5 & 78.7 & 79.9 & 85.4 & 55.3 & 98.0 & 12.3 \\
\rowcolor{blue!5} VisionZip (CVPR'25) \cite{yang2025visionzip} & 82.6 & 79.5 & 61.2 & 2306.5 & 78.3 & \textbf{80.5} & \textbf{86.1} & 55.2 & 98.0 & 1.1 \\
\rowcolor{green!5} DivPrune (CVPR'25) \cite{alvar2025divprune} & 82.3 & 78.6 & 61.4 & 2230.6 & 78.5 & 80.0 & 84.5 & 54.7 & 97.0 & 23.1 \\
\rowcolor{green!5} CDPruner (NeurIPS'25) \cite{zhang2025cdpruner} & 83.7 & 79.6 & 63.4 & 2302.2 & \textbf{79.2} & 80.3 & \textbf{86.1} & \textbf{55.9} & 98.8 & 112.4 \\
\rowcolor{green!5} VisionSelector \cite{VisionSelector} & \textbf{84.5} & 79.5 & \textbf{63.8} & 2315.8 & 77.4 & 79.9 & 85.8 & 53.3 & 98.1 & 1.2 \\
\rowcolor{green!5} D$^2$Pruner (AAAI'26) \cite{zhang2025d2pruner} & 83.2 & \textbf{82.9} & \textbf{63.7} & 2305.3 & 78.2 & 79.6 & 85.3 & 54.9 & 98.7 & 31.2 \\
\rowcolor{green!5} \textbf{SFPruner} (Ours) & 84.0 & 79.9 & 63.1 & \textbf{2319.3} & 79.0 & 79.7 & \textbf{86.5} & \textbf{55.9} & \textbf{99.0} & 2.5 \\

\midrule
\rowcolor{orange!8}
\multicolumn{11}{c}{\emph{Retain 20\% Tokens (Avg. 256 Tokens)}} \\
\rowcolor{blue!5} PruMerge (ICCV'25) \cite{shang2025prumerge} & 74.7 & 77.2 & 59.1 & 2280.1 & 77.6 & 77.7 & 84.5 & 52.8 & 94.6 & \textbf{1.1} \\
\rowcolor{blue!5} VisPruner (ICCV'25) \cite{vispruner} & 75.3 & 77.4 & 59.0 & 2213.9 & 77.5 & 78.3 & 84.5 & 52.2 & 94.3 & 7.5 \\
\rowcolor{blue!5} VisionZip (CVPR'25) \cite{yang2025visionzip} & 75.9 & 77.6 & 59.1 & 2276.5 & 77.4 & 78.5 & 84.5 & 53.5 & 95.1 & \textbf{1.1} \\
\rowcolor{green!5} DivPrune (CVPR'25) \cite{alvar2025divprune} & 76.5 & 76.6 & 58.1 & 2203.7 & 76.9 & 77.5 & 83.6 & 51.8 & 93.7 & 11.8 \\
\rowcolor{green!5} CDPruner (NeurIPS'25) \cite{zhang2025cdpruner} & 79.1 & 78.4 & \textbf{61.1} & 2264.5 & 77.9 & \textbf{79.5} & 84.9 & 53.4 & 96.2 & 56.7 \\
\rowcolor{green!5} VisionSelector \cite{VisionSelector} & \textbf{83.4} & 78.1 & 60.7 & 2283.4 & 76.2 & 77.1 & 84.6 & 51.2 & 95.7 & 1.1 \\
\rowcolor{green!5} D$^2$Pruner (AAAI'26) \cite{zhang2025d2pruner} & 78.6 & \textbf{79.4} & 60.2 & 2280.5 & 76.1 & 78.6 & 83.0 & 52.1 & 95.3 & 18.9 \\
\rowcolor{green!5} \textbf{SFPruner} (Ours) & 79.8 & 78.3 & 60.2 & \textbf{2295.7} & \textbf{78.1} & 79.4 & \textbf{85.7} & \textbf{53.7} & \textbf{96.5} & 2.5 \\
\bottomrule
\end{tabular}}
\vspace{1mm}
\caption{\textbf{Main Results on Qwen2.5-VL-7B.} Performance comparison across various token pruning methods at different retention ratios. Relative performance (\textbf{Rel. (\%)}) is normalized to the unpruned baseline. Light blue rows indicate heuristic-based methods, and light green rows represent optimization-based methods.}
\label{tab:main_results}
\vspace{-5mm}
\end{table*}

\subsection{Performance on Qwen2.5-VL: High-Resolution Single Sequence}

We further evaluate the proposed method on Qwen2.5-VL-7B to examine scalability under denser visual sequences. 
Unlike LLaVA-NeXT, which processes high-resolution images by dividing them into independent patches, 
Qwen2.5-VL encodes the entire image as a single continuous token sequence. 
This architectural setting increases the number of tokens competing within a single attention window 
and therefore presents a more demanding redundancy modeling scenario.

As shown in Table~\ref{tab:main_results}, 
optimization-based methods incur noticeable selection overhead 
under this dense single-sequence regime. 
At 40\% token retention (512 tokens), 
CDPruner requires 112.4\,ms, 
DivPrune requires 23.1\,ms, 
and D$^2$Pruner requires 31.2\,ms for subset selection. 
At 20\% retention (256 tokens), 
their latency decreases proportionally to the reduced target token count 
(e.g., CDPruner: 56.7\,ms; DivPrune: 11.8\,ms; D$^2$Pruner: 18.9\,ms), 
reflecting the iterative nature of their selection process.
In contrast, our method maintains an identical selection time of 2.5\,ms 
for both 40\% and 20\% retention settings. 
This retention-invariant behavior arises from the absence of iterative subset construction, 
since redundancy modeling is resolved through parallel covariance-level attenuation 
and ranking-based directional masking.

Despite the reduced computational overhead, 
the proposed framework preserves competitive reasoning performance. 
At 40\% retention, our method achieves 99.0\% relative performance, 
comparable to CDPruner (98.8\%) and D$^2$Pruner (98.7\%). 
Furthermore, compared with the recently proposed \emph{learned} VisionSelector, 
SFPruner consistently achieves higher relative performance at both 
40\% (99.0\% vs.\ 98.1\%) and 20\% (96.5\% vs.\ 95.7\%) retention, 
despite requiring no additional training. 
Under more aggressive compression (20\% retention), 
performance remains at 96.5\%, 
again comparable to optimization-based methods 
while requiring substantially lower selection latency.
These results demonstrate that the proposed non-iterative formulation 
remains stable under long continuous token sequences, 
achieving redundancy-aware pruning without the latency growth 
associated with sequential subset construction.

\begin{table*}[t]
\centering
\small
\vspace{1mm}
\resizebox{0.95\textwidth}{!}{
\renewcommand{\arraystretch}{1.1}
\begin{tabular}{lcccccccccc} 
\toprule
\multirow{2}{*}{\textbf{Method}} & \multicolumn{3}{c}{\textbf{Video-MME}} & \textbf{MLVU} & \multicolumn{3}{c}{\textbf{LongVideoBench}} & \multirow{2}{*}{\textbf{MMVBench}} & \multirow{2}{*}{\textbf{Rel. (\%)}} & \textbf{Time} \\
\cmidrule(lr){2-4} \cmidrule(lr){5-5} \cmidrule(lr){6-8}
& \textbf{Short} & \textbf{Medium} & \textbf{Long} & \textbf{Avg.} & \textbf{Val} & \textbf{Perception} & \textbf{Relation} & & & \textbf{(ms)} $\downarrow$ \\
\midrule
\rowcolor{orange!8}
\multicolumn{11}{c}{\emph{Vanilla 100\% Tokens}} \\
LLaVA-Video-7B & 67.1 & 62.8 & 47.6 & 60.3 & 61.1 & 65.0 & 53.8 & 58.8 & 100.0 & - \\
\midrule
\rowcolor{orange!8}
\multicolumn{11}{c}{\emph{Retain 512 Tokens per Frame}} \\
%PruMerge (CVPR'25) \cite{yang2025visionzip} & - & - & - & - & - & - & - & 58.2 & - & \textbf{1.1} \\
\rowcolor{blue!5} VisPruner (ICCV'25) \cite{vispruner} & 64.0 & 50.8 & 44.0 & 57.5 & 53.5 & 52.5 & 53.5 & 56.5 & 91.0 & 12.3 \\
\rowcolor{blue!5} VisionZip (CVPR'25) \cite{yang2025visionzip} & 65.0 & 51.5 & 44.5 & 58.5 & 54.2 & 53.0 & 54.0 & 58.1 & 92.4 & \textbf{1.1} \\
\rowcolor{green!5} DivPrune (CVPR'25) \cite{alvar2025divprune} & 66.6 & \textbf{53.0} & 45.0 & 60.1 & 55.6 & \textbf{55.2} & 55.5 & 58.1 & 94.5 & 131.0 \\
\rowcolor{green!5} CDPruner (NeurIPS'25) \cite{zhang2025cdpruner} & 64.6 & 52.0 & 45.7 & 60.0 & 54.9 & 54.9 & 54.7 & 57.1 & 93.5 & 92.0 \\
\rowcolor{green!5} D$^2$Pruner (AAAI'26) \cite{zhang2025d2pruner} & 65.5 & 52.5 & 46.0 & 59.5 & 55.0 & 54.5 & 54.8 & 57.5 & 93.8 & 31.2 \\
\rowcolor{green!5} \textbf{SFPruner} (Ours) & \textbf{67.4} & 52.6 & \textbf{46.0} & \textbf{60.2} & \textbf{55.7} & \textbf{55.2} & \textbf{55.6} & \textbf{58.3} & \textbf{94.9} & 3.1 \\
\midrule
\rowcolor{orange!8}
\multicolumn{11}{c}{\emph{Retain 256 Tokens per Frame}} \\
%PruMerge (CVPR'25) \cite{yang2025visionzip} & - & - & 45.9 & - & - & - & - & 57.4 & - & \textbf{1.1} \\
\rowcolor{blue!5} VisPruner (ICCV'25) \cite{vispruner} & 60.5 & 47.0 & 41.5 & 54.0 & 50.5 & 49.5 & 49.5 & 53.0 & 85.3 & 7.5 \\
\rowcolor{blue!5} VisionZip (CVPR'25) \cite{yang2025visionzip} & 61.5 & 48.0 & 42.0 & 55.0 & 51.0 & 50.5 & 50.0 & 54.5 & 86.8 & \textbf{1.1} \\
\rowcolor{green!5} DivPrune (CVPR'25) \cite{alvar2025divprune} & 64.1 & 51.2 & 45.3 & \textbf{58.3} & \textbf{54.1} & \textbf{56.3} & 52.6 & 54.9 & 91.9 & 64.0 \\
\rowcolor{green!5} CDPruner (NeurIPS'25) \cite{zhang2025cdpruner} & 61.1 & 49.6 & 46.0 & 57.2 & 52.4 & 52.9 & 51.9 & 52.8 & 89.4 & 44.0 \\
\rowcolor{green!5} D$^2$Pruner (AAAI'26) \cite{zhang2025d2pruner} & 62.1 & 49.9 & 45.4 & 57.5 & 52.8 & 52.0 & 52.2 & 55.2 & 90.0 & 18.9 \\
\rowcolor{green!5} \textbf{SFPruner} (Ours) & \textbf{64.1} & \textbf{51.8} & \textbf{47.0} & 58.2 & 53.1 & 53.2 & \textbf{53.0} & \textbf{56.7} & \textbf{92.1} & 3.1 \\
\bottomrule
\end{tabular}
}
\vspace{1mm}
\caption{\textbf{Main Results on Video MLLM Benchmarks.} Performance comparison using the LLaVA-Video-7B model across multiple frames. Relative performance (\textbf{Rel. (\%)}) is normalized to the unpruned baseline. Light blue rows indicate heuristic-based methods, and light green rows represent optimization-based methods.}
\label{tab:videomainresults}

\end{table*}

\subsection{Performance on Video Sequences: Multi-Frame Processing}

We further evaluate scalability under multi-frame settings using LLaVA-Video-7B. 
Video inference introduces substantially larger token counts due to temporal stacking, 
making redundancy modeling more challenging than in single-image settings. 
Table~\ref{tab:videomainresults} reports performance and selection latency 
under 512 and 256 tokens per frame.

Sequential optimization-based methods incur noticeable selection overhead in this regime. 
At 512 tokens per frame, DivPrune requires 131.0\,ms, 
CDPruner requires 92.0\,ms, and D$^2$Pruner requires 31.2\,ms. 
In contrast, our method maintains a stable selection latency of 3.1\,ms, 
comparable to lightweight heuristic approaches. 
Moreover, the latency remains unchanged when reducing the retention budget 
from 512 to 256 tokens per frame, 
indicating retention-invariant pruning overhead under batch processing.

Despite the reduced computational cost, 
the proposed framework preserves competitive spatio-temporal reasoning accuracy. 
At 512 tokens per frame, our method achieves 94.9\% relative performance, 
the highest among compared methods. 
Under stronger compression (256 tokens per frame), 
performance remains at 92.1\%, 
comparable to optimization-based baselines while maintaining significantly lower selection latency. 
These results demonstrate that the proposed non-iterative formulation 
remains stable under temporal batching, 
achieving redundancy-aware pruning without the latency growth 
associated with sequential subset construction.

\subsection{Extreme Scalability: High-Resolution Stress Test}

To further evaluate structural scalability under extreme conditions, we conduct a stress test by scaling up the input resolution to an ultra-high level on Qwen2.5-VL using the entire MME dataset \cite{mme},
producing a continuous sequence of 9,216 visual tokens ($D=3584$).
We apply a 30\% retention budget (approximately 2,770 tokens)
and profile peak GPU memory, isolated pruning time,
end-to-end prefill latency, and total inference time.
The results are summarized in Table~\ref{tab:extreme_scale}.
At this scale, sequential subset-optimization methods incur substantial pruning overhead.
For example, CDPruner requires 576\,ms for subset selection,
while DivPrune requires 458\,ms.
In contrast, the proposed method completes pruning in 28\,ms.
Importantly, the impact of pruning overhead becomes more pronounced
as token count increases.
In this 9,216-token regime,
optimization-based methods exhibit limited end-to-end speedup
due to selection overhead dominating attention savings.
By comparison, our non-iterative formulation reduces prefill latency
from 1813\,ms (vanilla) to 1114\,ms, and total inference time for the dataset from 6101\,sec to 3601\,sec,
demonstrating that token reduction translates into practical acceleration
when pruning overhead is bounded.

\begin{wraptable}{r}{0.62\textwidth} 
\vspace{-15pt}
\centering

\vspace{2pt}
\scriptsize 
\renewcommand{\arraystretch}{1.1}
\setlength{\tabcolsep}{4pt} 
\resizebox{0.62\textwidth}{!}{

\begin{tabular}{lccccc} 
\toprule
\textbf{Method} & \textbf{FLOPs} & \textbf{Peak Memory} & \textbf{Pruning} & \textbf{Prefill} & \textbf{Infer Time} \\
& (T) $\downarrow$ & (MB) $\downarrow$ & (ms) $\downarrow$ & (ms) $\downarrow$ & (sec) $\downarrow$ \\
\midrule
\rowcolor{orange!8}
\multicolumn{6}{c}{\emph{Vanilla 100\% Tokens}} \\
Full Seq & 148 & 19,749 & - & 1813 & 6101 \\
\midrule
\rowcolor{orange!8}
\multicolumn{6}{c}{\emph{Retain 30\% Tokens}} \\
DivPrune (CVPR'25)& 44.4  & 18,512 & 458 & 1547 & 5387 \\
CDPruner (NeurIPS'25)& 44.4 & 18,736 & 576 & 1658 & 5787 \\
%D$^2$Pruner (AAAI'26) & 18,595 & 132.4 & 912.3 \\
\textbf{SFPruner} (Ours) & 44.4 & \textbf{18,432} & \textbf{28} & \textbf{1114} & \textbf{3601} \\
\bottomrule
\end{tabular}}
\caption{\textbf{Hardware Profiling under Extreme Scale.} Qwen2.5-VL ($N=9,216$).}
\label{tab:extreme_scale}
\vspace{-15pt} 
\end{wraptable}
Peak memory usage remains comparable across methods.
Our approach consumes 18,432\,MB,
similar to other optimization-based baselines
and lower than the unpruned full sequence (19,749\,MB).
This confirms that covariance-level redundancy attenuation
does not introduce additional memory burden
even in ultra-high-resolution settings.

\subsection{Micro-Architectural Profiling and Code-Level Analysis}
\begin{figure*}[t]
    \centering
    \includegraphics[width=0.85\linewidth]{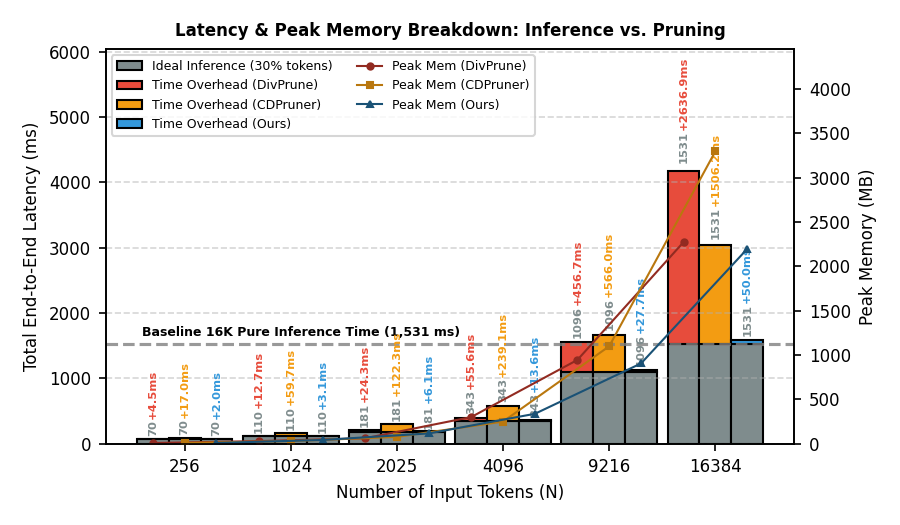}

    \caption{\textbf{End-to-end latency and peak memory analysis across varying numbers of input tokens ($N$).} While existing token pruning methods (DivPrune, CDPruner) incur severe time overheads as sequence length increases, ours maintains exceptionally low overhead even at a high resolution of $N=16384$. It closely approaches the pure inference time of the 16K baseline (1,531 ms), delivering practical computational acceleration. Bar charts indicate total latency, and line plots represent peak memory.}
    \label{fig:latency_breakdown}
    \vspace{-2mm}
\end{figure*}

To comprehend the limited speedups of existing methods at high resolutions, we analyzed pruning overheads. As illustrated in Fig.~\ref{fig:latency_breakdown}, the total latency is expressed as $(T_{\text{total}} = T_{\text{ideal inference}}(K) + T_{\text{pruning}}(N))$.

We conducted benchmarks across various token counts while maintaining previous settings. By replicating the extreme scaling scenario with $(N=9,216)$ tokens and a 30\% retention budget, we isolated the execution of the pruning module. Averaging over 10 independent trials with explicit \texttt{torch.cuda.synchronize()} boundaries, the pure selection latency consistently converged to 27.7 ms. This consistency between isolated function-level profiling and macroscopic measurements (e.g., Table 4) demonstrates similar outcomes across different token counts.
Methods like DivPrune and CDPruner employ sequential autoregressive loops repeated (K) times, resulting in inefficient GPU utilization. With (N=16,384), pruning overheads reach 2.6 seconds and 1.5 seconds, negating any inference speedup.

Our approach addresses this by eliminating (K)-dependent loops. As noted in Sec. 3.2, we utilize the Woodbury matrix identity when (N > D) to circumvent the costly $(\mathcal{O}(N^3))$ inversion of the $(N \times N)$ similarity matrix. This reduces the problem to a $(D \times D)$ feature covariance matrix, solved using Cholesky decomposition for speed and numerical stability, reducing computational complexity to $(\mathcal{O}(N D^2))$. As shown in Fig.~\ref{fig:latency_breakdown}, this optimization achieves substantial latency reduction without increasing memory usage; peak memory remains comparable to existing baselines. Consequently, our method executes pruning on 16,384 tokens in just 50.0 ms, translating theoretical FLOP reductions into real-world acceleration.
We rigorously validate our latency claims via detailed, function-level micro-benchmarks. Our parallel tensor formulation effectively circumvents the sequential bottlenecks encountered in previous subset-optimization techniques, such as iterative search loops. This approach enables stable execution without incurring additional kernel-launch overheads. Further experimental details can be found in the supplementary material.

\subsection{Ablation Study} \label{subsec:ablation}

We conduct ablation experiments on LLaVA-NeXT-7B, retaining 64 tokens per patch (320 tokens in total), to evaluate the contributions of key components in SFPruner. Table~\ref{tab:ablation_core} summarizes task-specific accuracy, relative performance retention (Rel.\%), and selection latency.

\paragraph{\textbf{Directional Masking vs. Sequential Selection.}} To assess the effect of pairwise redundancy control, we compare our ranking-based directional masking with a naive Top-$K$ baseline and a conventional sequential selection strategy, all employing the baseline semantic guidance (SG). As shown in Table~\ref{tab:ablation_core}, the naive Top-$K$ approach yields the lowest retention (96.9\%) despite its low latency (0.7 ms) due to unresolved spatial redundancy. Sequential selection achieves the highest retention among SG-based methods (98.0\%) by explicitly updating token states step-by-step, but suffers from substantial latency (18.6 ms) due to iterative computation. In contrast, directional masking operates as a parallel process, achieving a comparable retention (97.9\%) while drastically reducing latency to 0.8 ms. These results indicate that directional masking provides an efficient and effective means of mitigating local overlaps without incurring iterative bottlenecks.

\paragraph{\textbf{Effect of Covariance-Level Attenuation.}}
To isolate the contribution of covariance-level redundancy modeling, we replace the semantic guidance (SG) with the proposed Semantics-Guided Ridge Leverage Score (SG-RLS) while retaining the same Top-$K$ selection strategy. As shown in Table~\ref{tab:ablation_core}, SG-RLS consistently improves relative performance from 96.9\% to 97.6\%, demonstrating that covariance-level attenuation alone effectively suppresses globally redundant tokens. When combined with directional masking, the relative performance further increases to 98.3\% with consistent gains across tasks (e.g., GQA: 61.2 $\rightarrow$ 61.5, POPE: 86.4 $\rightarrow$ 86.7). Although covariance matrix operations introduce a modest overhead (+1.5\,ms), the total latency (2.3\,ms) remains substantially lower than that of sequential search (18.6\,ms). These results indicate that covariance-level attenuation and directional masking are complementary, addressing global and local redundancy, respectively.

\begin{table}[t]
\centering

\vspace{1mm}
\setlength{\tabcolsep}{8pt} 
\renewcommand{\arraystretch}{0.9}
\resizebox{0.95\textwidth}{!}{

\begin{tabular}{llccccc}
\toprule
\textbf{Scoring Metric} & \textbf{Selection Strategy} & \textbf{GQA} & \textbf{TextVQA} & \textbf{POPE} & \textbf{Rel. (\%)} & \textbf{Time (ms)} $\downarrow$ \\
\midrule
SG ($S_{\text{guide}}$) & Top-$K$ Selection & 60.3 & 57.8 & 85.5 & 96.9 & \textbf{0.7} \\
SG-RLS (Ours) & Top-$K$ Selection & 60.5 & 58.1 & 86.1 & 97.6 & 2.2 \\
SG ($S_{\text{guide}}$) & Sequential Search & 61.2 & 58.2 & 86.3 & 98.0 & 18.6 \\
SG ($S_{\text{guide}}$) & Directional Masking & 61.2 & \textbf{58.3} & 86.4 & 97.9 & 0.8 \\
\midrule
\rowcolor{orange!8}
\textbf{SG-RLS (Ours)} & \textbf{Directional Masking} & \textbf{61.5} & \textbf{58.3} & \textbf{86.7} & \textbf{98.3} & 2.3 \\
\bottomrule
\end{tabular}}
\vspace{1mm}
\caption{\textbf{Ablation on Scoring Metrics and Selection Strategies.} Evaluated on LLaVA-NeXT-7B retaining 64 tokens per patch (320 tokens in total). The synergistic combination of SG-RLS and ranking-based directional masking achieves the optimal balance.}
\label{tab:ablation_core}

\end{table}

\section{Conclusion}
\label{sec:conclusion}

This paper revisited the computational behavior of token pruning in high-resolution Multimodal Large Language Models (MLLMs). We observed that while subset-optimization-based methods provide strong redundancy control, their iterative selection procedures introduce non-negligible overhead in dense visual regimes. Motivated by this observation, we proposed SFPruner, a non-iterative token pruning framework that performs redundancy-aware importance selection through structured tensor operations. SFPruner decomposes redundancy modeling into two complementary components. Semantics-Guided Ridge Leverage (SG-RLS) attenuates covariance-level redundancy by rebalancing dominant feature directions, while ranking-based directional masking resolves residual pairwise overlap without sequential subset construction. This formulation enables structured redundancy control while avoiding iterative dependency in the pruning process. Extensive experiments across diverse MLLM architectures, including multi-patch high-resolution (LLaVA-NeXT), single-sequence dense encoding (Qwen2.5-VL), and multi-frame video processing (LLaVA-Video), demonstrate that SFPruner consistently converts token reduction into practical latency improvement. Across evaluated settings, it maintains competitive reasoning performance while significantly reducing selection overhead. Overall, the results suggest that redundancy-aware token pruning can be realized without iterative subset construction, providing a scalable alternative for high-resolution and long-context multimodal inference.

\section*{Acknowledgements}

% \textbf{Do not} include acknowledgements in the initial version of the paper

This work was supported by the National Research Foundation of Korea (NRF) grant funded by the Korean government (MSIT) (No. RS-2024-00456152). This work was also supported by LG Electronics. Computational resources were provided by “the Advanced GPU Utilization Support Program” funded by the Government of the Republic of Korea (Ministry of Science and ICT) and the Cluster Server for Computational Science at Pusan National University.

\bibliographystyle{splncs04}
\bibliography{main}

\end{document}